\documentclass[5p,authoryear]{elsarticle} 

\usepackage{framed,multirow}
\usepackage{amssymb}
\usepackage{latexsym}
\usepackage{xcolor}
\usepackage{lineno,hyperref}
\modulolinenumbers[5]
\usepackage{amsmath}
\usepackage{multirow}
\usepackage{color,graphicx}
\usepackage{nicematrix}
\usepackage{arydshln}
\usepackage{hyperref}
\usepackage{url}
\usepackage{footnote}
\usepackage{bm}
\usepackage{cite}
\usepackage{subfigure}
\usepackage{verbatim}
\usepackage{ulem}
\usepackage{graphicx}
\usepackage{pgfplots}
\usepackage{cancel}

\usepackage{booktabs}  
\usepackage{tabularx}

\usepackage{enumitem}

\journal{Medical Image Analysis}

\begin{document}

\begin{frontmatter}

\title{Personalized 4D Whole-Heart Mesh Reconstruction from Cine MRI via Multi-Scale Temporal Modeling and Differentiable Contour Rendering}

\author[label1]{Xiaoyue Liu} 
\author[label1]{Dongcheng Cang} 
\author[label1,label2]{Xiaohan Yuan} 
\author[label3,label4]{Mark YY Chan} 
\author[label3,label4]{Ching-Hui Sia} 
\author[label1]{Lei Li*} 
\ead[url]{lei.li@nus.edu.sg}

\address[label1]{Department of Biomedical Engineering, National University of Singapore, Singapore}
\address[label2]{School of Automation, Southeast University, Nanjing, China}
\address[label3]{Department of Medicine, National University of Singapore, Singapore}
\address[label4]{Department of Cardiology, National University Heart Centre Singapore, Singapore}

\begin{abstract}
Accurate 4D whole-heart mesh reconstruction from sparse cine MRI is critical for creating cardiac digital twins, but remains challenging due to limited 2D slice coverage and the complex coupling between cardiac shape and motion. Existing methods often rely on intermediate contour fitting and typically reconstruct static, single-phase, or partial cardiac geometries, limiting their ability to capture full-chamber dynamics. We propose a novel end-to-end framework for reconstructing temporally resolved whole-heart meshes from multi-view 2D cine MRI sequences by learning an image-to-mesh mapping. The framework incorporates a differentiable contour renderer inspired by the Beer-Lambert attenuation principle, enabling anatomy-aware supervision of 3D+t mesh deformation through contour-based projection losses. To improve temporal consistency across the cardiac cycle, we further introduce a multi-scale temporal modeling module that integrates global cycle-level dynamics with local inter-frame coherence to generate smooth and physiologically plausible mesh trajectories. The proposed method achieved a whole-heart mean absolute error of 1.68 $\pm$ 0.31 mm and a motion jitter of 0.77 $\pm$ 0.17 $\mathrm{mm}/\mathrm{frame}^{3}$, outperforming existing methods with lower reconstruction error and substantially improved motion smoothness. It also improved 2D contour alignment across multiple cine MRI views and supported downstream proof-of-concept electrophysiological simulation. The code will be released publicly upon acceptance of the manuscript for publication.

\end{abstract}

\begin{keyword}
4D Whole Heart Reconstruction \sep Cine MRI \sep Multi-Scale Temporal Modeling \sep Differentiable Rendering \sep Cardiac Digital Twin
\end{keyword}

\end{frontmatter}

\section{Introduction}

Cardiovascular diseases remain the leading cause of mortality worldwide, making reliable assessment of cardiac structure and function essential for clinical diagnosis and therapeutic decision-making \citep{journal/EJPC/chong2025}. 
In this context, cardiac digital twin (CDT) technology has emerged as a powerful tool for creating patient-specific virtual heart models, enabling real-time visualization and analysis of the heart \citep{journal/EBME/li2024}.
By offering detailed insights into the underlying mechanisms of the heart, CDT has the potential to revolutionize cardiac diagnosis and treatment \citep{journal/EHJ/corral2020,journal/NC/arevalo2016}.
A typical CDT workflow consists of two key stages: anatomical twinning and functional twinning \citep{journal/MedIA/gillette2021,journal/TMI/li2024}.
Anatomical twinning involves extracting 3D heart geometry from images and identifying pathological regions when present. 
Considering the dynamic nature of cardiac motion, 4D (3D+t) geometry is typically required for a more comprehensive representation.
Cine MRI can be used for this purpose, as it provides non-invasive visualization of cardiac anatomy and motion throughout the cardiac cycle. 
However, cine MRI typically acquires sparse and intersecting 2D image planes, i.e., short-axis (SAX) and long-axis (LAX) slices, leaving substantial spatial gaps that make accurate 4D cardiac geometry reconstruction highly challenging. 

Conventional cine MRI based cardiac geometry reconstruction frameworks therefore generally consist of two steps, i.e., image segmentation, followed by mesh generation based on contours derived from the segmentation \citep{journal/MedIA/gaggion2025}. 
This is mainly because direct volumetric reconstruction from cine MRIs is challenging due to the inherent sparsity and anisotropy of the data.  
By first segmenting the cardiac structures, the extracted contours can serve as geometric constraints to guide the mesh generation. 
Nevertheless, traditional iso-surfacing algorithms, such as marching cubes, struggle to generate smooth and anatomically accurate meshes due to the irregular spacing and insufficient volumetric information in the input data.
To solve this, many previous studies employed mesh adaptation approaches, such as template mesh deformation \citep{journal/JI/villard2018,journal/CMPB/hu2023}, statistical shape model (SSM) \citep{journal/PTRSA/banerjee2021}, and B-spline surface reconstruction \citep{journal/MRM/odille2018}. 
Furthermore, image interpolation based methods have also been applied to reconstruct high-resolution 3D geometry \citep{journal/TMI/ukwatta2015}. 
However, these techniques are labor-intensive and time-consuming, which significantly hinders their feasibility for real-time applications.

Recently, deep learning based methods have shown promising performance for efficient 3D cardiac geometry reconstruction \citep{conf/ICCV/yuan2023,journal/MedIA/gaggion2025}.
Similar to the conventional pipeline, these deep learning based methods generally rely on pre-generated segmentation and then directly convert sparse contours into 3D meshes via point/ shape completion network \citep{journal/MedIA/beetz2023,conf/MICCAI/ma2025}, flow matching \citep{journal/arxiv/ma2026}, or graph convolution network based template deformation \citep{journal/MedIA/chen2021,conf/ICCV/ye2023}. 
Furthermore, most studies focus only on partial cardiac anatomy, such as the left ventricular (LV) myocardium \citep{journal/MedIA/joyce2022,conf/ICCV/yuan2023,journal/TMI/meng2023} or the biventricular model \citep{journal/MedIA/chen2021,journal/MedIA/beetz2023}, reducing their applicability to whole-heart functional twinning. 
Other approaches that achieve whole-heart reconstruction rely on high-resolution volumetric CT or MRI data \citep{journal/TMI/kong2022}, which are not part of standard clinical workflows. 
Most critically, recent frameworks that reconstruct from sparse 2D cine MRI primarily operate on static 3D shape at isolated cardiac phases \citep{journal/MedIA/gaggion2025,conf/MICCAI/ma2025}.
They failed to model the continuous temporal dynamics and the intrinsic coupling of cardiac shape and motion throughout the cycle. 
This omission limits the physiological fidelity required for dynamic simulations.
In general, these work either rely on high-resolution volumetric images or solely reconstructed part of the whole heart or single phase of the heart.

In this study, we introduce a novel framework that reconstructs temporally resolved whole-heart meshes directly from sparse multi-view cine MRI by jointly learning shape and motion information. 
Given multi-view cine MRIs, the model can efficiently learn an image-to-mesh mapping between cine MRI representations and mesh deformation space.
Specifically, we utilize the domain-specific autoencoder networks to extract the compact latent representations of both cine MRIs and the mesh sequences. 
Then, the mapping between the image and heart mesh sequences latent spaces can be learned to ensure the generated shapes align with the mesh deformation space. 
This study is a systematic extension of our previous conference work \citep{conf/MICCAI/liu2026}. 
We provide more comprehensive quantitative functional evaluation, including quantitative
assessment of ejection fraction and chamber volume.
We also include external validation on the public dataset to assess cross-domain generalizability. 
More importantly, we demonstrate the feasibility of using the reconstructed whole-heart meshes as anatomical substrates for downstream in-silico electrophysiological (EP) simulation.  
The main contributions of this work include:
\begin{enumerate}[label=\roman*.]
  \item We present a novel end-to-end image-to-mesh pipeline that reconstructs patient-specific 4D whole-heart meshes directly from multi-view 2D cine MRI sequences.
  \item We design a physics-inspired differentiable rendering loss, based on the Beer–Lambert law, which leverages multi-view 2D+t segmentation contours as direct supervision for 3D+t mesh reconstruction. 
  \item We develop a multi-scale temporal modeling module that models high-dimensional image-sequence dynamics to produce a smooth and physiologically plausible latent trajectory for 4D mesh generation. 
  \item We demonstrate the feasibility of using reconstructed whole-heart meshes for proof-of-concept EP simulation, supporting their potential integration into CDT workflows.
\end{enumerate}

\section{Related Work} 

\subsection{3D Cardiac Reconstruction from Sparse 2D Medical Imaging}

Reconstructing patient-specific 3D anatomy from sparse 2D observations is a persistent problem in cardiac image analysis across different modalities including echocardiography (echo), X-ray angiography, and MRI \citep{journal/VCIBA/sarmah2023}. 
In echo, limited acoustic windows yield only a few intersecting 2D planes, from which the complete 3D cardiac anatomy must be inferred despite poor boundary visibility and speckle artifact \citep{journal/MedIA/laumer2025,conf/MICCAI/yu2025}.
In biplane X-ray angiography, two near-orthogonal projections are used to reconstruct coronary vessel geometry, relying on epipolar constraints and vesselness priors to resolve depth ambiguity \citep{conf/ISBI/bransby2023}. 
In sparse-slice MRI, where only a small number of slices are acquired, reconstructing complete anatomical surfaces requires strong regularization to fill unsampled regions \citep{journal/TMI/ukwatta2015}. 
Despite differences in imaging physics, these tasks share the challenge of reconstructing complete 3D anatomy from partial, view-dependent observations.

Traditional approaches for reconstructing cardiac anatomy from sparse 2D cine MRI generally follow a segmentation-to-shape pipeline, where SAX and LAX images are first converted into intermediate contours or shape observations and then used to fit statistical shape models \citep{journal/TMI/ukwatta2015}, deform template meshes \citep{journal/MRM/odille2018}, or construct spline-based surfaces \citep{journal/PTRSA/banerjee2021}. Although these methods provide explicit geometric constraints, their reconstruction quality depends heavily on hand-crafted shape fitting. More recent learning-based methods have introduced data-driven components into this pipeline, including volume super-resolution \citep{journal/TMI/oktay2017}, point-to-mesh prediction \citet{conf/ICCV/ye2023}, and shape deformation \citep{journal/TMI/xiao2024,journal/MedIA/chen2021}. However, many of these methods still rely on intermediate segmentations, contours, regular-grid representations, or point-cloud inputs, rather than directly reconstructing complete anatomical meshes from sparse multi-view cine MRI.
Moreover, most approaches only focus on one or two ventricular structures \citep{journal/MedIA/chen2021,conf/ISBI/beetz2021}, providing an incomplete representation of the whole heart. Also, many learning based models rely on simulated SAX and LAX MRI slices from 3D images for training, which may limit their adaptability to real 2D cine MRI \citep{conf/ISBI/beetz2021,journal/MedIA/chen2021,conf/STACOM/xu2023}.

\begin{figure*}[t]\center
 \includegraphics[width=1\textwidth]{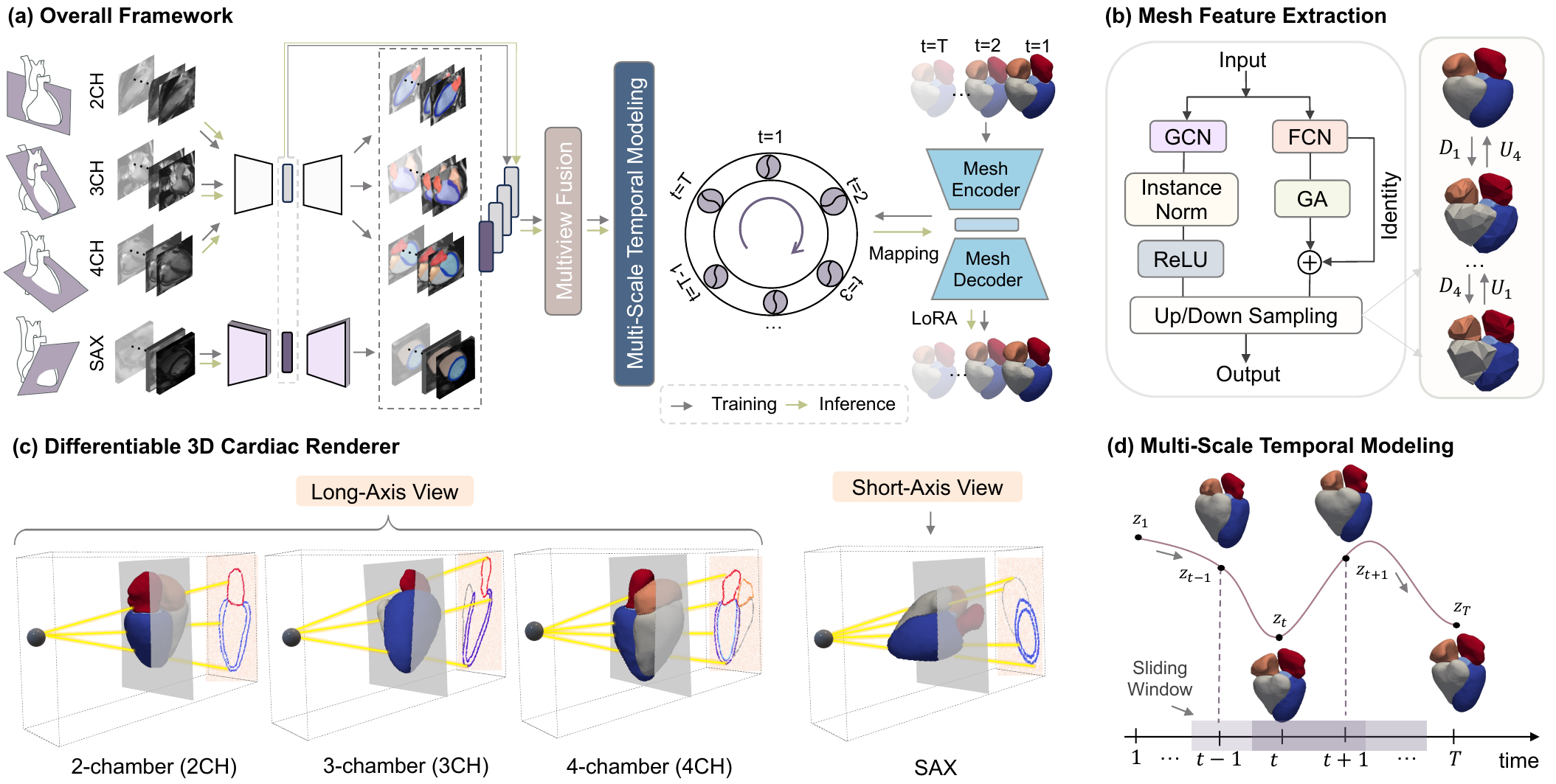}\\[-2ex]
   \caption{Illustration of the proposed multi-view cine-MRI based 4D whole heart mesh reconstruction framework. GCN: graph convolutional network; 2/ 3/ 4CH: 2/ 3/ 4-chamber; SAX: short-axis; FCN: fully convolutional network; GA: graph attention; LoRA: low-rank adaptation. }
\label{fig:method:framework}
\end{figure*}

\subsection{Cardiac Motion Modeling and Analysis}
Cardiac motion analysis is fundamental for evaluating cardiac function and characterizing abnormal mechanical behavior \citep{journal/MedIA/yu2014}. 
Recovering myocardial deformation and chamber dynamics enables quantitative assessment of ejection fraction \citep{journal/Nature/ouyang2020}, strain \citep{journal/MedIA/lopez2023}, and other clinically relevant biomarkers that support diagnosis, risk stratification, and treatment planning \citep{journal/NatureMI/bello2019}.
Different imaging modalities offer complementary strengths for motion analysis. 
Echo provides high temporal resolution and broad clinical accessibility, with motion estimation typically performed by tracking speckle or intensity patterns, though challenges remain in out-of-plane motion handling and signal dropout \citep{journal/TMI/evain2022, journal/Nature/ouyang2020}. 
Tagged MRI embeds explicit magnetic tag patterns into the myocardium, allowing precise tracking of tissue displacement and strain across the cardiac cycle, though its specialized acquisition limits routine use \citep{journal/MedIA/wang2019}. 
In routine clinical practice, cine MRI is more commonly available than tagged MRI, offering dynamic anatomical information without specialized sequences \citep{journal/NatureMI/bello2019,journal/MedIA/lopez2023}. 


Depending on the representation, cine-based methods may estimate motion at the image, point cloud, or mesh level. 
At the image level, cine MRI motion tracking is commonly formulated as dense frame-to-frame or full-cycle motion estimation, either through registration-based alignment or optical-flow-based displacement recovery. 
Traditional registration methods recover cine MRI motion by spatially aligning image frames and estimating deformation fields under regularization or physical constraints \citep{journal/MedIA/bistoquet2008,journal/MedPhys/qiao2020}. 
Optical-flow-based methods estimate cine MRI motion as dense pixel-wise displacement fields from local image correspondences or learned apparent-flow representations \citep{journal/MedIA/wang2019,journal/MedIA/zheng2019}.
Although image-level cine motion tracking can estimate dense pixel-wise displacement fields, it does not directly provide explicit anatomical surface trajectories or mesh-based geometric measurements such as vertex-wise correspondence, chamber-wise deformation, and time-resolved surface geometry \citep{conf/MICCAI/meng2022,journal/TMI/meng2023}.
At the point-cloud level, cine-based cardiac motion modeling represents reconstructed cardiac surfaces as unordered 3D point sets and estimates anatomical deformation between cardiac phases. 
For example, point-cloud deformation has been used to model biventricular contraction and relaxation between end-diastolic and end-systolic phases from cine-derived cardiac point clouds \citep{journal/JBHI/beetz2024}.
However, because point clouds are unordered, they lack explicit surface connectivity and fixed vertex correspondence, which limits their direct use for topology-preserving cardiac surface motion tracking \citep{journal/CAG/lemeunier2022}.
Among these, mesh-based motion reconstruction is particularly attractive because it preserves anatomical correspondence over time and enables direct analysis of chamber deformation, myocardial motion, and subject-specific functional biomarkers \citep{journal/arxiv/lyu2026,conf/MICCAI/yang2026}.
Single-chamber mesh modeling has been explored by \citet{journal/MedIA/joyce2022}, who fitted personalized LV meshes to sparse cine MRI slices throughout the cardiac cycle to derive functional measurements specific to the LV such as volume and strain.
A representative biventricular approach is DeepMesh, which estimates 3D cardiac mesh motion from multi-view cine MRI and preserves vertex correspondence across frames for ventricular motion analysis \citep{journal/TMI/meng2023}. 
However, single- or biventricular mesh tracking only provides a partial representation of cardiac dynamics, limiting its ability to capture coordinated motion across the whole heart. 
Recently, \citet{journal/arxiv/ma2026} attempted to synthesis whole-heart cardiac motion using flow matching.
Such synthesis-based methods mainly generate plausible motion patterns rather than reconstructing patient-specific anatomical dynamics directly constrained by cine MRI observations.

\subsection{Whole-Heart Simulation for Cardiac Digital Twins}
Digital twins have emerged as a powerful paradigm for personalized medicine, enabling in silico prediction of patient-specific cardiac function, treatment response, and disease progression \citep{journal/EHJ/corral2020,journal/NC/arevalo2016}. 
However, the majority of existing cardiac simulation studies have focused on isolated chambers, most commonly the ventricles, reflecting the relative maturity and clinical importance of ventricular modeling \citep{journal/MedIA/gillette2021,journal/TMI/li2024,journal/MedIA/li2025}. 
Atrial-only simulations, while increasingly studied in the context of arrhythmias such as atrial fibrillation, typically neglect atrioventricular coupling and the mechanical influence of ventricular contraction on atrial dynamics \citep{journal/MedIA/zappon2025,journal/NatureBME/boyle2019}. 
Conversely, ventricular-only models often ignore the atria altogether, treating them as passive pressure boundary conditions or omitting them entirely. 
This compartmentalized approach fails to capture physiologically important atrioventricular interactions  \citep{journal/CMAME/fedele2023}.
The gap persists because the atria exhibit thin-walled, complex trabeculated anatomy and are physiologically coupled to ventricular dynamics, making their inclusion computationally demanding and data-intensive. 
Furthermore, personalizing whole-heart models requires anatomical reconstructions of all four chambers, tissue property estimation across heterogeneous structures, and validation against multi-chamber functional data, all of which remain nontrivial. 
Consequently, advancing whole-heart digital twins to fully capture atrioventricular interaction remains an essential and still-evolving frontier in personalized cardiac modeling.
Recently, several studies have taken important steps toward whole-heart simulation. 
For instance, \citet{journal/CMAME/fedele2023} developed a biophysically detailed electromechanical model of the whole human heart that considers both atrial and ventricular contraction, incorporating patient-specific geometry, fiber architecture, and tissue properties to simulate coupled electromechanics across all four chambers. 
In parallel, \citet{journal/FiP/gillette2022} proposed a personalized virtual model of entire organ-scale electrophysiology of all four-chambers of the heart.
More recently, \citet{journal/npj/salvador2024} introduced a data-driven framework using latent neural ordinary differential equations to accelerate whole-heart electromechanical simulations, achieving substantial speedups while preserving biophysical fidelity.
Despite these advances, whole-heart simulation remains computationally expensive and sensitive to parameter choices. 
Most existing methods rely on high-resolution volumetric imaging to construct detailed patient-specific meshes, and they typically assume the availability of dense 3D geometry at a single representative time point, with motion either prescribed or simulated from first principles rather than directly inferred from routine clinical acquisitions. 

\section{Methodology}

Figure~\ref{fig:method:framework} (a) provides an overview of the proposed 4D whole heart reconstruction model, consisting of domain-specific feature extractors and latent dynamic mapping.
To achieve 4D mesh reconstruction from cine MRI, we design a cross-domain mapping module to bridge the image domain and mesh domain (Sec.~\ref{method:mapping}).
U-Net and mesh VAE networks are employed to extract the compact latent representations of cine MRIs and mesh sequences, respectively.
Additionally, a multi-scale temporal modeling module is introduced to capture cycle-level cardiac dynamics and inter-frame coherence (Sec.~\ref{method:temporal_modeling}).
Then, we employ a differentiable rendering to explicitly learn the spatial relationship among the 2D(+t) sparse cardiac segmentation and 3D(+t) mesh (Sec.~\ref{method:rendering}). 
Finally, Sec.~\ref{method:inference} presents the details of the reconstruction model for the personalized inference of 4D whole heart mesh.

\subsection{Image-to-Mesh Latent Space Mapping via Anatomical Feature Alignment} \label{method:mapping}

For image feature extraction, we employ a pretrained cardiac U-Net \citep{conf/MICCAI/ronneberger2015} to extract the anatomy of cardiac chambers. Given multi-view cine sequences $\{x_t^{2CH}, x_t^{3CH}, x_t^{4CH}, x_t^{SAX}\}_{t=1}^{T}$, the U-Net generates pixel-wise chamber segmentation masks $\{S_t\}_{t=1}^{T}$ that explicitly delineate anatomical boundaries.
The optimization of U-Net is supervised using Dice score loss between predicted and manually labeled cardiac regions:
\begin{equation}
\label{eq:mesh_vae_loss}
\mathcal{L}_{\mathrm{U-Net}} = 1 - Dice(S_t, \hat{S}_t).
\end{equation}
The obtained anatomical feature embeddings $\mathbf{z}_t^{\text{anatomy}}$ from different views can be fused as the cardiac anatomical representation for downstream image-mesh mapping. 

For mesh feature extraction, we use a pretrained mesh VAE to learn a comprehensive prior over both cardiac geometry and temporal dynamics. 
Each 4D mesh sequence $\{M_t\}_{t=1}^{T}$ is represented as a sequence of spatial graphs
$\mathcal{G}_t=(\mathcal{V}_t,\mathcal{E})$, where
$\mathcal{V}_t$ denotes time-dependent vertex coordinates and
$\mathcal{E}$ denotes fixed mesh connectivity.
The mesh VAE employs a hierarchical encoder-decoder architecture, as presented in Fig.~\ref{fig:method:framework} (b). 
A spatial graph convolutional network (GCN) branch captures local geometric structure, while a parallel graph attention branch, implemented via Exphormer \citep{conf/ICML/shirzad2023}, models long-range dependencies. 
The outputs of these branches are fused through a residual skip connection. The encoder comprises a series of mesh feature extraction blocks, each followed by downsampling to progressively reduce spatial resolution while expanding feature dimensions. 
Fixed downsampling matrices ($\mathbf{D}_m$), precomputed on the atlas surface mesh using the method in \citet{conf/ECCV/ranjan2018}, transition among different resolution levels.
At the bottleneck, the encoder maps the input mesh at time step $t$ to parameters of a latent Gaussian distribution, i.e., $\mathbf{z}_t^{\text{mesh}}$. 
The decoder mirrors this architecture, using upsampling matrices ($\mathbf{U}_m$) to reconstruct the mesh from the latent sample.
The mesh VAE is optimized by minimizing:
\begin{equation}
\label{eq:mesh_vae_loss}
\mathcal{L}_{\mathrm{MeshVAE}} = \frac{1}{T} \sum_{t=1}^{T} \left\| M_t - \hat{M}_t \right\|^2 + \lambda_{KL} \, \mathcal{R}_{\mathrm{KL}},
\end{equation}
where $M_t$ and $\hat{M}_t$ represent the input and reconstructed mesh at time $t$, and $\mathcal{R}_{\mathrm{KL}}$ is the Kullback-Leibler (KL) divergence term. 

The core of our approach is to learn an optimal mapping $\mathcal{T}_{I\to M}$ between the anatomical embedding space $\mathbf{z}^{\text{anatomy}}$ and the geometric embedding space $\mathbf{z}^{\text{mesh}}$. 
For efficient adaptation, we use low-rank adaptation (LoRA) \citep{conf/ICLR/hu2022} applied to the decoder of mesh VAE (see Sec. \ref{method:inference}). 
This design enables personalized 4D mesh reconstruction by mapping segmentation-based anatomical features to geometrically plausible mesh sequences through the learned latent correspondence, while maintaining the rich geometric priors encoded in the frozen mesh VAE components.

\begin{figure}[t]\center
 \includegraphics[width=0.49\textwidth]{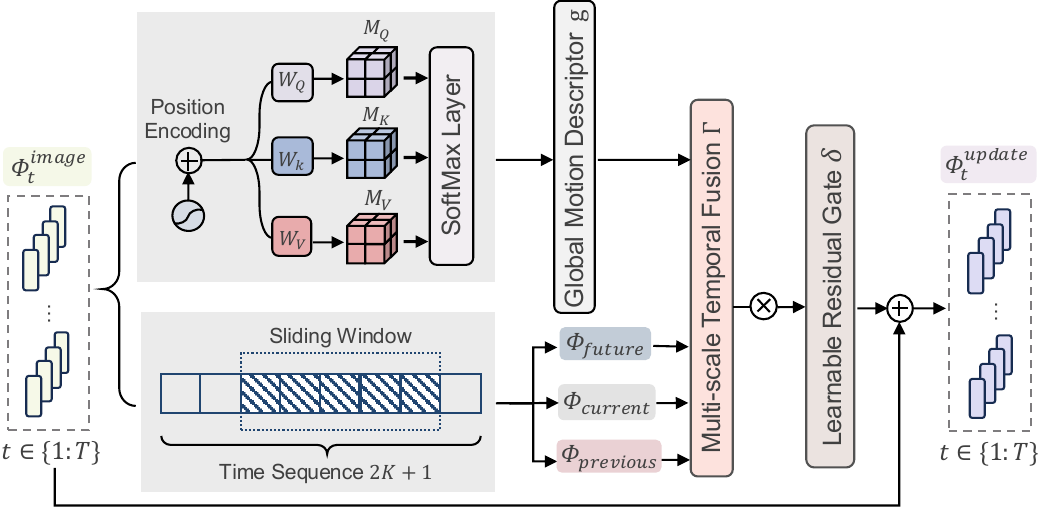}\\[-2ex]
   \caption{Multi-scale temporal modeling on the fused image features. }
\label{fig:method:temporal_modeling}
\end{figure}

\subsection{Multi-Scale Latent Dynamics Modeling for Temporal Coherence} \label{method:temporal_modeling}

Modeling clinically meaningful cardiac function requires capturing both cycle-level temporal patterns (global feature) and short-term inter-frame consistency (local feature). To address this requirement, we propose a multi-scale temporal (MST) modeling module, as shown in Fig.~\ref{fig:method:framework} (d). This framework operates on fused anatomical embeddings and explicitly models temporal evolution at multiple scales.
At each time step \( t \), the anatomical latent embeddings \(\mathbf{z}_t^{\mathrm{anatomy}}\) from multiple imaging views (LAX and SAX) are fused via a multi-view module consisting of fully-connected layers and linear projection, yielding a unified representation \(\Phi_t^{\mathrm{image}}\). To preserve temporal ordering, sinusoidal positional encoding \citep{conf/NeuIPS/vaswani2017} is added to the fused latent sequence.

As illustrated in Fig.~\ref{fig:method:temporal_modeling}, our temporal modeling architecture employs two complementary branches. A global branch applies temporal self-attention to the stacked latent embeddings, with pooled outputs yielding a global motion descriptor \(\mathbf{g}\) that encodes sequence-wide temporal context. Simultaneously, a local branch enforces short-term consistency among neighboring frames through a sliding window mechanism. For each reference frame \( t \), neighboring frames within a temporal sliding window \(\mathcal{W}_t\) of total size \((2K+1)\) are grouped into three relative temporal categories: past frames \(\Phi_{\mathrm{previous}}=\{\Phi_{t-j}^{\mathrm{image}}\}_{j=1}^{K}\), the current frame \(\Phi_{\mathrm{current}}=\Phi_{t}^{\mathrm{image}}\), and future frames \(\Phi_{\mathrm{future}}=\{\Phi_{t+j}^{\mathrm{image}}\}_{j=1}^{K}\). This grouping enables modeling of direction-aware temporal dependencies.
We define \(\Gamma\) as a multi-scale fusion module that jointly leverages the global features \(\mathbf{g}\) and the local neighborhood window \(\mathcal{W}_t\) to produce a temporally informed latent representation. This is implemented by feature-space stacking of sequence-level and sliding-window context, fused by a compact projection head. The complete temporal update is formulated as:
\begin{equation}
\hat{\Phi}_t = \Gamma(\mathbf{g},\mathcal{W}_t),\;
\Phi_t^{\mathrm{update}} = \Phi_t^{\mathrm{image}} + \delta \odot \hat{\Phi}_t, 
\end{equation}
where \(\delta\) is a learnable residual gating vector that controls the integration of temporal information into the geometry-aware latent state.

\subsection{Differentiable Rendering for Contour-Guided Mesh Optimization} \label{method:rendering}

Multi-view cine MRI captures cardiac anatomy through sparse view-specific planes, providing complementary yet inherently 2D observations of a 3D heart. To reconstruct a complete 4D mesh under such planar supervision, we require a differentiable coupling that translates vertex-to-plane distances into continuous per-view contributions, enabling stable contour-driven mesh refinement.
Our approach is motivated by the physical principle of exponential attenuation described in the Beer-Lambert law \citep{journal/ChemPhys/mayerhofer2020}, where transmitted light intensity diminishes exponentially with increasing path length through a medium. This provides a natural computational analogy: mesh vertices positioned near an imaging plane should exert stronger influence on the projected contour, while distant vertices contribute negligibly, with the influence decaying smoothly as distance increases.

As illustrated in Fig.~\ref{fig:method:framework} (c), for each anatomical view $x \in \{\mathrm{2CH}, \mathrm{3CH}, \mathrm{4CH}, \mathrm{SAX}\}$, we extract the affine transformation from the fixed image header to recover the view-plane geometry $\Pi^{\,x}$ in world coordinates. The predicted surface-mesh vertices $\hat{\mathbf{v}}_t^{\,i}$ at time frame $t$ are represented in the same coordinate system, where $i$ indexes individual mesh vertices.
Each imaging plane $\Pi^{\,x}$ is parameterized by a point $\mathbf{c}^{\,x}$ lying on the plane and a unit normal vector $\mathbf{n}^{\,x}$, both defined in world coordinates. The perpendicular distance from a predicted vertex to this plane is computed as:
\begin{equation}
D_t^{\,i,x} = \mathrm{dist}\!\left(\hat{\mathbf{v}}_t^{\,i},\,\Pi^{\,x}\right) = \left|\mathbf{n}^{\,x}\cdot\left(\hat{\mathbf{v}}_t^{\,i}-\mathbf{c}^{\,x}\right)\right|.
\label{eq:vertex_plane_distance}
\end{equation}
Drawing upon the Beer-Lambert analogy, we transform these distances into plane-association probabilities $p_t^{\,i,x}$ that quantify how strongly each vertex contributes to the projected view. A sigmoid-weighted distance function $\ell_{\mathrm{sigmoid}}(\cdot)$ combined with exponential attenuation yields:
\begin{equation}
p_t^{\,i,x} = 1 - \exp\!\left(-\mu\,\ell_{\mathrm{sigmoid}}(D_t^{\,i,x})\right),
\label{eq:plane_prob}
\end{equation}
where $\ell_{\mathrm{sigmoid}}(D)=\sigma\!\left(\alpha(\tau-D)\right)$ defines a decreasing proximity window with respect to the vertex-to-plane distance.
Here, $\alpha$ controls the steepness of the sigmoid transition, $\tau$ determines the effective slab thickness around the imaging plane, and $\mu$ controls the saturation strength of the exponential attenuation. 
Vertices close to the imaging plane therefore obtain higher association probabilities, whereas geometrically distant vertices are smoothly down-weighted. 
The sigmoid window provides a smooth differentiable transition around the imaging plane, while the exponential term controls how strongly this proximity response is converted into a plane-association probability.
These vertex-wise probabilities are then projected onto their respective view planes and aggregated to construct probability maps $P_t^{\,x}$. This aggregation mechanism ensures that vertices in close proximity to the imaging plane contribute prominently, while geometrically distant vertices are progressively suppressed, naturally emphasizing the anatomical contours visible in each slice.
The differentiable rendering (DR) objective is formulated as:
\begin{equation}
\mathcal{L}_{\mathrm{DR}} = \sum_{x} \mathcal{L}_{\mathrm{boundary}}(P_t^{\,x}, S_t^{\,x}),
\label{eq:dr_loss}
\end{equation}
where $\mathcal{L}_{\mathrm{boundary}}$ implements the boundary constraint proposed in \citet{journal/MIA/kervadec2019}, and $S_t^{\,x}$ represents the ground-truth segmentation mask for view $x$ at time $t$. This loss is computed across all temporal frames, providing dense supervisory signals throughout the cardiac cycle. The boundary loss formulation is particularly appropriate here, as it operates on contour distances rather than requiring pixel-wise classification, aligning naturally with our contour-based supervision paradigm.


\subsection{Personalized Inference of 4D Whole-Heart Mesh} \label{method:inference}

During inference, the image encoder $f_e^{\mathrm{I}}$ is kept frozen. The learned image-to-mesh mapping $\mathcal{T}_{I\to M}$, which encodes MST features, serves as a conditioning signal to the mesh decoder $f_d^{\mathrm{M}}$ for generating subject-specific, temporally coherent 4D whole-heart meshes. The generation process is formulated as:
\begin{equation}
\label{eq:inference}
\hat{M} =
\mathrm{LoRA}\!\left(f_{d}^{\mathrm{M}}\right)
\Big(
\mathcal{T}_{I\to M} \big(
f_{e}^{\mathrm{I}}(I)
\big)
\Big),
\end{equation}
where $\mathrm{LoRA}(\cdot)$ denotes the low-rank adaptation blocks integrated into the linear layers of the mesh decoder to enable parameter-efficient adaptation under a fixed backbone \citep{conf/ICLR/hu2022}.
The two modality-specific feature extractors ($f^{\mathrm{I}}$ for image and $f^{\mathrm{M}
}$ for mesh) are trained independently. To ensure geometric regularity of the predicted surface, we employ an edge-length regularizer $\mathcal{L}_{\mathrm{edge}}$ and a normal-consistency regularizer $\mathcal{L}_{\mathrm{norm}}$.
$\mathcal{L}_{\mathrm{edge}}$ discourages abnormal stretching or shrinkage by penalizing variations in edge lengths, while $\mathcal{L}_{\mathrm{norm}}$ stabilizes local surface orientation by regularizing face normals \citep{conf/ECCV/wang2018}.
The reconstruction fidelity is measured by the mean squared error $\mathcal{L}_{\mathrm{MSE}}$ between the predicted mesh $\hat{M}$ and the ground-truth mesh $M$. Combined with the regularizer $\mathcal{L}_{\mathrm{DR}}$ defined in Eq.~\eqref{eq:dr_loss}, the total optimization objective is:
\begin{equation}
\label{eq:total_loss}
\begin{aligned}
\mathcal{L}_{\text{total}} = &\,
\lambda_{\mathrm{MSE}}\,\mathcal{L}_{\mathrm{MSE}}(\hat{M},\, M) 
+ \lambda_{\mathrm{DR}}\,\mathcal{L}_{\mathrm{DR}} \\
&+ \lambda_{\mathrm{edge}}\,\mathcal{L}_{\mathrm{edge}}
+ \lambda_{\mathrm{norm}}\,\mathcal{L}_{\mathrm{norm}},
\end{aligned}
\end{equation}
where $\lambda_{\mathrm{MSE}}$, $\lambda_{\mathrm{DR}}$, $\lambda_{\mathrm{edge}}$, and $\lambda_{\mathrm{norm}}$ are weighting coefficients balancing the respective loss terms.

\section{Experiments and Results} 

\subsection{Data Acquisition and Pre-Processing} \label{exp:data}

We collected 222 multi-view cine MRI scans from post myocardial infarction (MI) patients at the National University Hospital (NUH) in Singapore. 
Each patient has up to two sets of scans, with a time interval of approximately 6 months to 12 months.
Specifically, the set includes a stack of SAX balanced steady-state free precession (bSSFP) cine MRI and three LAX cine sequences (2-, 3-, and 4-chamber views), acquired using Siemens 3T scanner. 
The SAX cine sequences consist of 7 to 14 slices across 25 frames.
All images were cropped into a unified size of 150 $\times$ 150 centering at the heart region, with intensity normalization via Z-score. 
The dataset was randomly divided into 155 training, 10 validation, and 67 test samples.
To avoid data leakage, all data from a single patient were kept within the same data split (training, validation, or testing).

\begin{table*}[t] \center
\caption{Quantitative results of whole-heart and cardiac substructure mesh reconstruction. LV: left ventricle; RV: right ventricle; LA: left atrium; RA: right atrium; MAE: mean absolute error; MSE: mean squared error.}
{\footnotesize 
\begin{tabular}{l l c c c c c}
\hline
 & & Whole Heart & LV & RV & LA & RA \\
\hline
\multirow{3}{*}{HybridVNet} & MAE (mm) $\downarrow$ & 2.18 ± 0.54 &  1.70 ± 0.54 & 1.97 ± 0.55 & 2.31 ± 0.70 & 2.51 ± 0.60 \\
& MSE (mm$^2$) $\downarrow$ & 8.80 ± 5.31 &  5.10 ± 3.67 & 7.02 ± 4.72 & 9.58 ± 9.24 & 11.72 ± 10.14 \\
& ${J}_{{m}}$ ($\mathrm{mm}/\mathrm{frame}^{3}$) $\downarrow$ & 2.29 ± 0.23 & 1.73 ± 0.16 & 2.07 ± 0.15 & 2.27 ± 0.25 & 2.70 ± 0.28\\
\hline
\multirow{3}{*}{Ours} & MAE (mm) $\downarrow$ & 1.68 ± 0.31 & 1.59 ± 0.34 & 1.64 ± 0.35 & 1.99 ± 0.48 & 1.86 ± 0.58 \\
& MSE (mm$^2$) $\downarrow$ & 5.06 ± 1.79 & 4.45 ± 1.93 & 4.86 ± 2.15 & 6.90 ± 3.19 & 6.26 ± 4.06 \\
& ${J}_{{m}}$ ($\mathrm{mm}/\mathrm{frame}^{3}$) $\downarrow$  & 0.77 ± 0.17 & 0.69 ± 0.15 & 0.78 ± 0.17 & 0.91 ± 0.21 & 0.93 ± 0.23\\
\hline
\end{tabular} } \\
\label{exp:tb:substructure_comparison}
\end{table*}

\begin{table*}[t!]
\centering
\caption{Summary of 2D contour results of predicted mesh on different views. DR: differentiable rendering; MCD: Mean Contour Distance; BF: Boundary F-score.}
{
\fontsize{8}{11}\selectfont
\begin{tabular}{l cc cc cc}
\hline
 & \multicolumn{2}{c}{HybridVNet} & \multicolumn{2}{c}{Ours (w/o $\mathcal{L}_{\mathrm{DR}}$)} & \multicolumn{2}{c}{\textbf{Ours}} \\
\cline{2-7}
View & BF (\%) $\uparrow$ & MCD (mm) $\downarrow$ & BF (\%) $\uparrow$ & MCD (mm) $\downarrow$ & BF (\%) $\uparrow$ & MCD (mm) $\downarrow$ \\
\hline
2CH & 55.80 ± 12.05 & 2.85 ± 4.76 & 60.13 ± 9.89 & 2.30 ± 0.47 &\textbf{65.47 ± 10.41} & \textbf{1.99 ± 0.41} \\
3CH & 54.62 ± 9.01 & 3.60 ± 0.66 & 58.27 ± 9.31 & 2.32 ± 0.46 & \textbf{65.71 ± 10.32} & \textbf{1.97 ± 0.44} \\
4CH & 57.01 ± 9.14 & 3.69 ± 5.10 & 59.64 ± 8.68 & 2.42 ± 0.46 & \textbf{68.24 ± 9.24} & \textbf{1.89 ± 0.38}\\
SAX & 62.74 ± 13.56 & 2.22 ± 0.89 & 60.74 ± 11.53 & 2.29 ± 0.59 & \textbf{62.86 ± 13.74} & \textbf{2.03 ± 0.62} \\
\hline
\end{tabular}}
\label{exp:tb:view_comparison}
\end{table*}

\subsection{Ground Truth and Evaluation Metrics} \label{exp:label}

Cine MRI segmentation was first derived from an automated pipeline \citep{journal/sensor/tasmurzayev2025} followed by manual refinement to ensure segmentation quality.
The manual refinement was performed by a well-trained biomedical engineering student using ITK-SNAP and double checked by a senior expert. 
These refined segmentations were subsequently resampled into an aligned world coordinate system to form a sparse 3D volume, which is fed into the atlas-deformation process introduced by \citet{conf/STACOM/xu2024} to produce a dense 3D segmentation label map. 
By leveraging the fixed affine correspondence between the volume space and the surface mesh vertices of atlas, the dense label map was projected onto a unified high-resolution template to generate topology-consistent whole-heart surface meshes. 
Each mesh consists 88,424 vertices and 176,828 faces and includes five anatomical components, namely LV, LV myocardium (Myo), right ventricle (RV), left atrium (LA), and right atrium (RA).
This surface mesh was then served as the ground truth whole heart mesh.

For evaluation, we employed vertex-wise Mean Absolute Error (MAE) and Mean Squared Error (MSE) computed between the reference and predicted whole-heart surface meshes. 
Chamfer Distance (CD) and Hausdorff Distance (HD) were additionally reported for shape similarity assessment, and per-mesh inference time was measured for computational comparison. 
Furthermore, Mean Contour Distance (MCD) and boundary F-score (BF) were reported between the reference and predicted contours for each view, where MCD measures the average contour distance and BF quantifies boundary alignment \citep{conf/CVPR/cheng2019,conf/arxiv/gur2019end}.
${E}_{{vol}}$ was reported as the whole-heart volumetric error (mL), averaged over frames, using the absolute volume difference.
Mesh jitter (${J}_{m}$) measured the temporal smoothness of vertex trajectories over time \citep{conf/CVPR/shin2024,conf/CVPR/yi2022}.
For each 4D heart mesh data, chamber volumes were computed over time and the ejection fraction (EF) correlation has also been used as a measurement.

\subsection{Implementation Details}

The framework was implemented in PyTorch, running on a computer with a 13th Gen Intel(R) Core(TM) i9-13980HX CPU and an NVIDIA GeForce RTX 4090D GPU.
We used the Adam optimizer to update the network parameters via stochastic gradient decent. 
The balancing parameters in Sec.~\ref{method:inference} are set as follows: $\lambda_{\mathrm{MSE}} = 10$, $\lambda_{\mathrm{DR}} = 5$, $\lambda_{\mathrm{edge}} = 0.8$, and $\lambda_{\mathrm{norm}} = 0.8$.
We pretrained the MeshVAE for 350 epochs using a learning rate of $1 \times 10^{-4}$.
For image–mesh mapping, we trained the model for 400 epochs using the Adam optimizer with a learning rate of $5 \times 10^{-5}$ and a batch size of 4.
The temporal sliding window size is set to 5 with $K = 2$.
The learning rate for LoRA applied to the pretrained mesh decoder is set to 1e-5.


\begin{figure*}[t]
  \centering
  \includegraphics[width=0.9\linewidth]{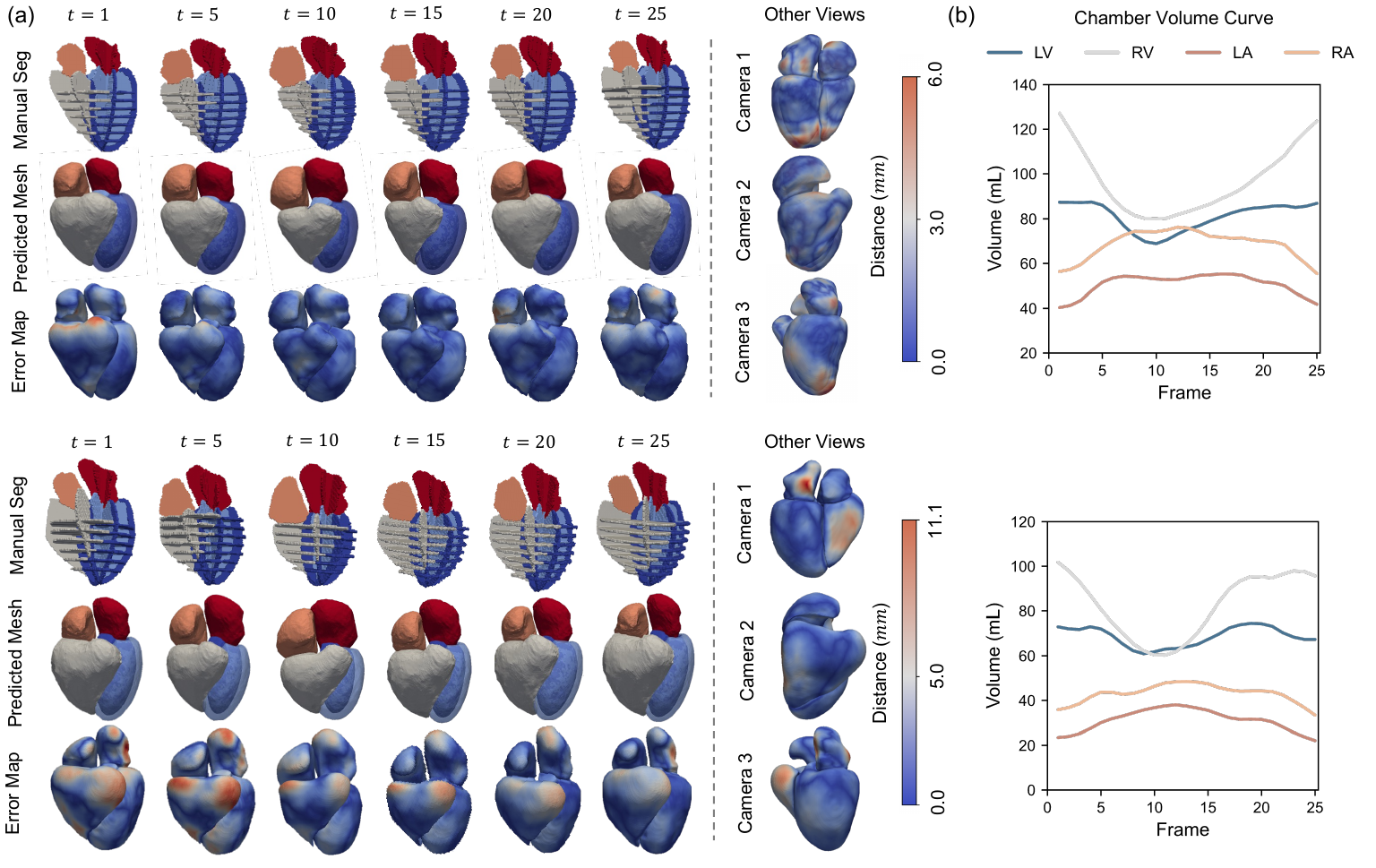}
  \caption{Illustration of whole-heart reconstruction quality of two representative cases. (a) Reconstructed 3D whole heart geometry at different cardiac phases and views. (b) Corresponding volume curves of reconstructed cardiac chambers.}
\label{fig:result:error_map}
\end{figure*}

\begin{figure*}[t]
  \centering
\includegraphics[width=0.96\linewidth]{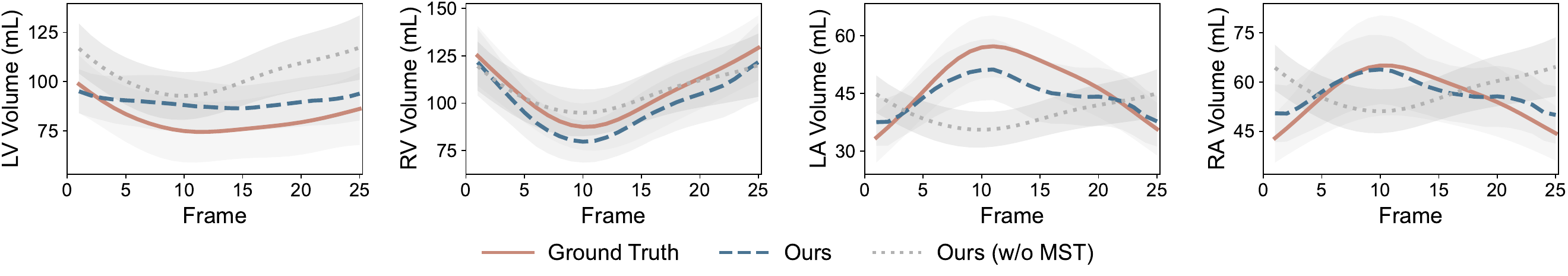}
  \caption{Illustration of chamber volume curves (mean $\pm$ std) of ground truth and predicted whole heart mesh. MST: multi-scale temporal.}
  \label{fig:result:volume}
\end{figure*}

\begin{figure*}[t]
    \centering
    \includegraphics[width=0.65\linewidth]{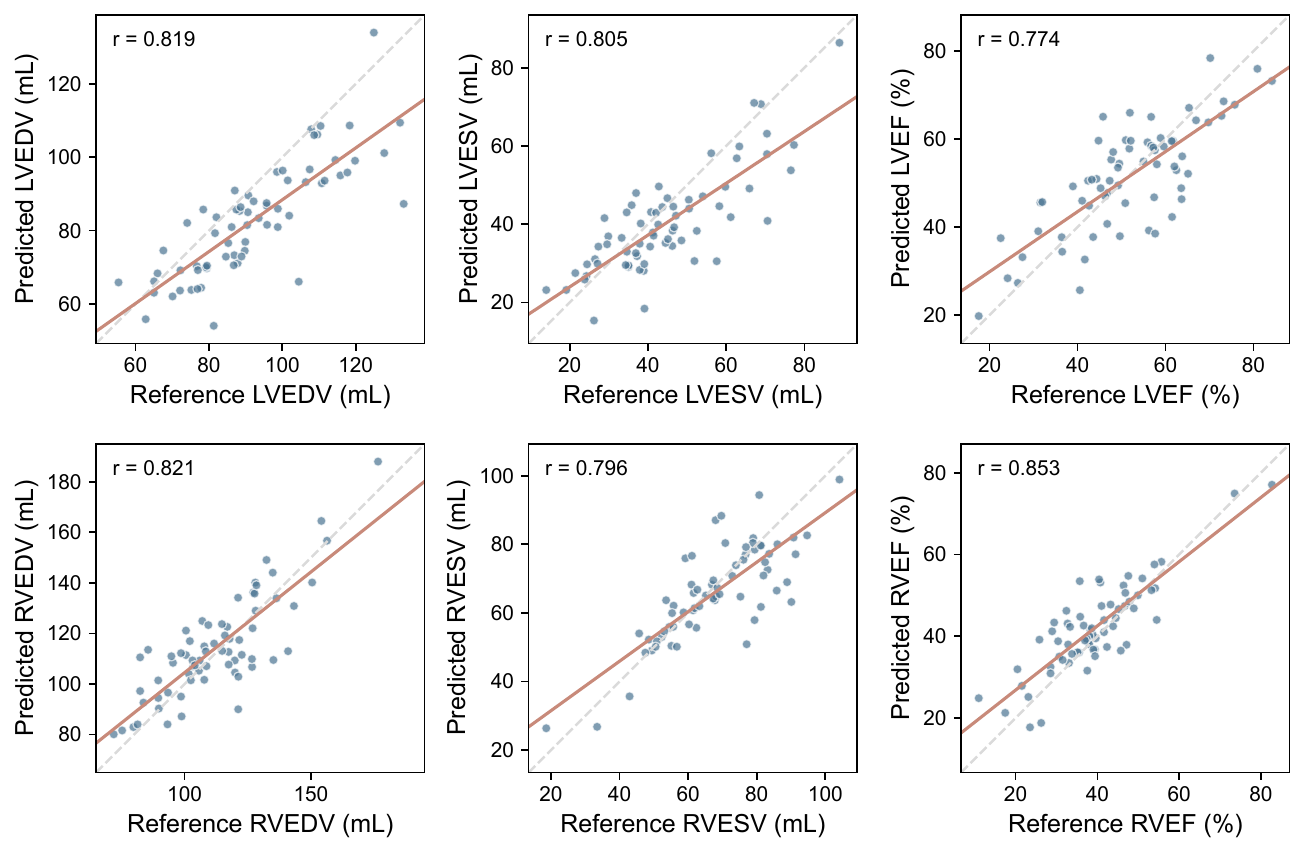}
    \caption{
    Correlation between predicted and reference ventricular functional indices. Pearson correlation coefficient ($r$) is reported in each panel. Dashed grey lines indicate identity, and solid red lines indicate linear regression fits. LVEDV/ RVEDV: LV/ RV end-diastolic volume; LVESV/ RVESV: LV/ RV end-systolic volume; LVEF/ RVEF: LV/RV ejection fraction. 
    }
    \label{fig:result:clinical_correlation}
\end{figure*}

\subsection{Results of the Proposed Method} \label{exp:result:ours}

The proposed method obtained a promising result for reconstructing 4D whole heart from cine MRI, with an average MAE of 1.68 ± 0.31 mm and MSE of 5.06 ± 1.79 mm$^{2}$.
Besides the overall performance of the proposed model on the whole heart, we also analyzed its performance across all cardiac substructures, as presented in Table \ref{exp:tb:substructure_comparison}.
One can see that compared to the baseline model, i.e., HybridVNet \citep{journal/MedIA/gaggion2025}, our model consistently obtained better performance at both global and chamber-specific levels.
Specifically, the left and right ventricles yield an average MAE of 1.59 ± 0.34 mm and 1.64 ± 0.35 mm, with corresponding MSE of 4.45 ± 1.93 mm$^{2}$ and 4.86 ± 2.15 mm$^{2}$.
For the left and right atria, average MAE are 1.99 ± 0.48 mm and 1.86 ± 0.58 mm, while average MSE are 6.90 ± 3.19 mm$^{2}$ and 6.26 ± 4.06 mm$^{2}$, respectively.
In general, we can observe that ventricular reconstructions achieved higher geometric accuracy than atrial reconstructions. 
This can be attributed to the fundamental asymmetry in cross-sectional sampling inherent to standard multi-view cardiac MRI protocols.
The ventricles are sampled through several short‑axis stacks, providing multiple, spatially correlated 2D contours along the long‑axis direction. 
In contrast, the atria are represented by only a single long‑axis slice in each imaging plane, resulting in a substantially sparser set of cross‑sectional observations. 
Consequently, the model has substantially more spatial information from which to infer the ventricular geometry, whereas atrial shape estimation must rely on far more limited and often incomplete anatomical evidence. 

This sampling-induced performance disparity is further corroborated by our 2D contour alignment analysis across different imaging views, as summarized in Table \ref{exp:tb:view_comparison}.
We observe that the SAX view, which comprises multiple slices along the ventricular long axis, achieves the lowest mean contour distance (2.03 ± 0.62 mm) among all views, indicating superior geometric alignment precision.
However, the SAX view yields a boundary F-score of 62.86 ± 13.74\%, which is slightly lower than those of the long-axis views (2CH: 65.47 ± 10.41\%, 3CH: 65.71 ± 10.32\%, 4CH: 68.24 ± 9.24\%).
This discrepancy can be explained by the inherent characteristics of each metric: mean contour distance measures the average Euclidean deviation between predicted and ground-truth contours, where SAX benefits from dense slice sampling that constrains the geometric solution space.
In contrast, boundary F-score balances precision and recall, and the lower score of SAX view may reflect the increased complexity of matching multiple parallel slices, where small misalignments across the stack can accumulate and reduce the overall boundary overlap.
The long-axis views, despite having sparser sampling, capture the chamber boundaries in a single sagittal plane, yielding higher boundary consistency at the expense of geometric precision.
These complementary observations collectively underscore the trade-offs inherent in multi-view reconstruction, where different imaging planes offer distinct strengths in capturing anatomical features.

Figure~\ref{fig:result:error_map} shows qualitative reconstruction results for two representative cases with different performance levels. As shown in Fig.~\ref{fig:result:error_map} (a), the proposed model reconstructed temporally coherent whole-heart meshes across the cardiac cycle from sparse cine MRI observations. In the first case, the predicted meshes closely followed the manual segmentations, with low surface errors across most cardiac structures and only minor local deviations near anatomically complex regions. In the second, more challenging case, larger errors were observed, particularly around the atrial and basal regions, reflecting the difficulty of recovering complete whole-heart anatomy from limited image evidence. Nevertheless, the reconstructed meshes remained anatomically plausible and temporally smooth across cardiac phases. The visualized error maps from additional viewpoints further indicated that the model preserved overall 3D structural consistency despite local reconstruction errors. Fig.~\ref{fig:result:error_map} (b) presents the corresponding chamber volume curves calculated based on predicted mesh, where the predicted temporal volume changes exhibited physiologically reasonable trends over the cardiac cycle.

Figure \ref{fig:result:volume} illustrates the temporal evolution of average chamber volumes derived from predicted and ground-truth meshes across the cardiac cycle. 
The predicted volumetric curves exhibit strong agreement with the ground truth in both magnitude and phase, demonstrating the ability of the proposed model to capture cardiac shape dynamics. 
However, the predicted LV volume change appears attenuated compared to the reference.
This observation is consistent with the pathological profile of our cohort, which comprises exclusively post-MI patients. 
In such patients, infarcted myocardial segments exhibit little to no contractile function, leading to a characteristically blunted global LV volume variation. 
Modeling this altered, pathology-specific motion pattern presents a distinct challenge compared to modeling healthy cardiac dynamics.
Notably, despite the inherent challenges posed by sparse cross-sectional sampling, the model successfully captures volume dynamics for the atrial chambers, which typically present greater reconstruction difficulty due to limited imaging evidence. 
We further examined the agreement between predicted and reference ventricular functional indices referring to \citet{journal/MedIA/gaggion2025} , as presented in Fig.~\ref{fig:result:clinical_correlation}. 
The reconstructed 4D whole-heart meshes achieved strong correlations for ventricular volumes, with $r=0.819$ for LV ED volume (LVEDV), $r=0.805$ for LV end-systolic volume (LVESV), $r=0.821$ for RV ED volume (RVEDV), and $r=0.796$ for RV ES volume (RVESV). 
Ejection fraction estimation also showed good agreement, with $r=0.774$ for LVEF and $r=0.853$ for RVEF. 
The relatively low LVEF and RVEF values observed in several cases are clinically plausible, as this is a post-MI cohort with potentially impaired ventricular contractility. 
Overall, these results indicate that the reconstructed 4D whole-heart meshes preserved clinically relevant volumetric dynamics and can provide reliable functional measurements.

\begin{table*}[t] 
\centering
\caption{Summary of the quantitative evaluation results of 4D whole heart mesh reconstruction of different methods.
Comparison results were reported in \citet{journal/MedIA/chen2021} and \citet{journal/MedIA/gaggion2025}. CD: Chamfer distance; HD: Hausdorff distance.}
{\footnotesize 
\begin{tabular}{lccc}
\hline
Method & CD (mm) $\downarrow$ & HD (mm) $\downarrow$ & Inference Time (s) $\downarrow$ \\
\hline
PointNet+ \citep{conf/NeurIP/qi2017}          & 13.03 ± 2.96 & 17.04 ± 3.57 & $<0.1$ \\
PU-Net \citep{conf/CVPR/yu2018}               & 12.15 ± 2.88 & 15.74 ± 3.37 & $<0.1$ \\
Pixel2mesh \citep{conf/ECCV/wang2018}         & 19.38 ± 5.54 & 16.20 ± 3.30 & $<0.1$ \\
CPD \citep{journal/TPAMI/myronenko2010}       & 12.10 ± 6.63 & 13.05 ± 7.04 & 37.45 \\
GMMREG \citep{journal/TPAMI/jian2010}         & 20.90 ± 7.18 & 18.57 ± 3.04 & 60.90 \\
MR-Net \citep{journal/MedIA/chen2021}         & 4.39 ± 1.48  & 6.89 ± 1.88  & $<0.1$ \\
HybridVNet \citep{journal/MedIA/gaggion2025}  & 4.13 ± 1.16  & 5.17 ± 1.02  & $<0.1$ \\
\hline
\textbf{Ours}                                 & \textbf{3.41 ± 0.62} & \textbf{5.13 ± 0.80} & $<0.1$ \\
\hline
\end{tabular}} 
\label{exp:tb:comparison_study}
\end{table*}


\subsection{Comparison Study} \label{exp:comparison}

We compare our method against representative point-based and mesh-based reconstruction approaches, with quantitative results summarized in Table~\ref{exp:tb:comparison_study}. Among all methods, our proposed model achieved the lowest Chamfer distance (3.41 ± 0.62 mm) and Hausdorff distance (5.13 ± 0.80 mm), outperforming the previous state-of-the-art HybridVNet \citep{journal/MedIA/gaggion2025} by margins of 0.72 mm and 0.04 mm, respectively. 
Notably, while traditional registration methods such as CPD~\citep{journal/TPAMI/myronenko2010} and GMMREG~\citep{journal/TPAMI/jian2010} require substantial computation time (up to 60.9 seconds per case), our method maintained real-time inference ($<0.1$ seconds) comparable to learning-based approaches. 
The superior performance of our method can be attributed to the synergistic integration of multi-view differentiable rendering and MST modeling, which together enforce both geometric precision and temporal coherence throughout the cardiac cycle.

We further analyze reconstruction performance across cardiac substructures and imaging views, as presented in Tables~\ref{exp:tb:substructure_comparison} and~\ref{exp:tb:view_comparison}. 
Compared to HybridVNet, our method achieved consistent improvements across all chambers, with whole-heart MAE decreasing from 2.18 ± 0.54 mm to 1.68 ± 0.31 mm (23.0\% reduction) and MSE from 8.80 ± 5.31 mm\(^2\) to 5.06 ± 1.79 mm\(^2\) (42.5\% reduction). 
The temporal smoothness metric \(J_m\) also improves substantially from 2.29 ± 0.23 to 0.77 ± 0.17 mm/frame\(^3\), confirming that our multi-scale temporal modeling yields physiologically plausible motion. 
Ventricular reconstructions remained more accurate than atrial ones across both methods. 
The differentiable rendering loss yields the most pronounced gains in long-axis views (2CH, 3CH, 4CH), with boundary F-score improvements of 17.33\%, 20.30\%, and 19.70\% relative to HybridVNet, respectively. 
These larger gains stem from the single-slice nature of long-axis views, where even small 3D deviations translate directly into contour misalignment, making direct boundary supervision particularly impactful. 
Qualitatively, the chamber volume curves in Fig.~\ref{fig:result:volume} demonstrated strong alignment with ground truth, capturing expected cardiac motion patterns and further validating the physiological consistency of our predictions.

\begin{table}[t]
\centering
\caption{Quantitative results of ablation studies evaluating different components of the proposed method for the 4D whole heart reconstruction. VAE: variational autoencoder; MST: multi-scale temporal. }
\label{exp:tb:ablation_study}
{\footnotesize  
\begin{tabular}{lccc}
\hline
Method & MSE $\downarrow$ & MAE $\downarrow$ & ${E}_{{vol}}$ $\downarrow$ \\
\hline
w/o SAX    & 5.64 ± 2.00 & 1.76 ± 0.31 & 25.3 ± 10.6 \\
w/o 2CH    & 5.59 ± 1.94 & 1.72 ± 0.29 & 27.3 ± 12.8 \\
w/o 3CH    & 5.87 ± 1.96 & 1.80 ± 0.30 & 25.3 ± 11.7 \\
w/o 4CH    & 5.99 ± 1.91 & 1.82 ± 0.28 & 26.3 ± 9.8  \\
w/o U-Net  & 6.43 ± 1.94 & 1.87 ± 0.28 & 30.8 ± 14.5 \\
w/o VAE    & 5.67 ± 2.03 & 1.77 ± 0.32 & 25.0 ± 10.8 \\
w/o MST     & 5.32 ± 2.02 & 1.69 ± 0.29 & 24.7 ± 13.0 \\
\textbf{Ours} & \textbf{5.06 ± 1.79} & \textbf{1.68 ± 0.31} & \textbf{17.3 ± 11.2} \\
\hline
\end{tabular}}
\end{table}

\subsection{Ablation Study} \label{exp:ablation_study}

We conduct ablation experiments to evaluate the contribution of each core component in our framework, with quantitative results presented in Table~\ref{exp:tb:ablation_study}. The full model achieves the lowest MSE (5.06 ± 1.79 mm\(^2\)), MAE (1.68 ± 0.31 mm), and volumetric error \(E_{\mathrm{vol}}\) (17.3 ± 11.2\%), demonstrating the synergistic effect of all components.
Multi-view feature fusion proves essential for accurate reconstruction. 
Removing any single view increases reconstruction error, with the 4CH view being most critical (MSE increases to 5.99 ± 1.91 mm\(^2\)), as it simultaneously captures all four chambers and provides comprehensive anatomical context. 
The SAX view contributes through-plane resolution critical for ventricular geometry, while long-axis views capture apical and longitudinal anatomy, together enabling complete 3D reconstruction. 
This confirms that complementary information from all imaging planes is necessary for accurate whole-heart modeling. 
Geometric priors from MeshVAE pretraining (w/o VAE) increase MSE to 5.67 ± 2.03 mm\(^2\) and substantially degrade volumetric error (25.0 ± 10.8\% versus 17.3 ± 11.2\% for the full model), indicating that constraining reconstructions to a plausible anatomical manifold benefits both spatial accuracy and temporal consistency. 
Segmentation-based pretraining via U-Net (w/o U-Net) yields the largest performance drop among all components (MSE: 6.43 ± 1.94 mm\(^2\)), as anatomical boundary features learned during segmentation provide critical edge-specific information for accurate mesh-image alignment, outperforming intensity-based self-supervision.

\begin{figure}[t]
  \centering
\includegraphics[width=0.9\linewidth]{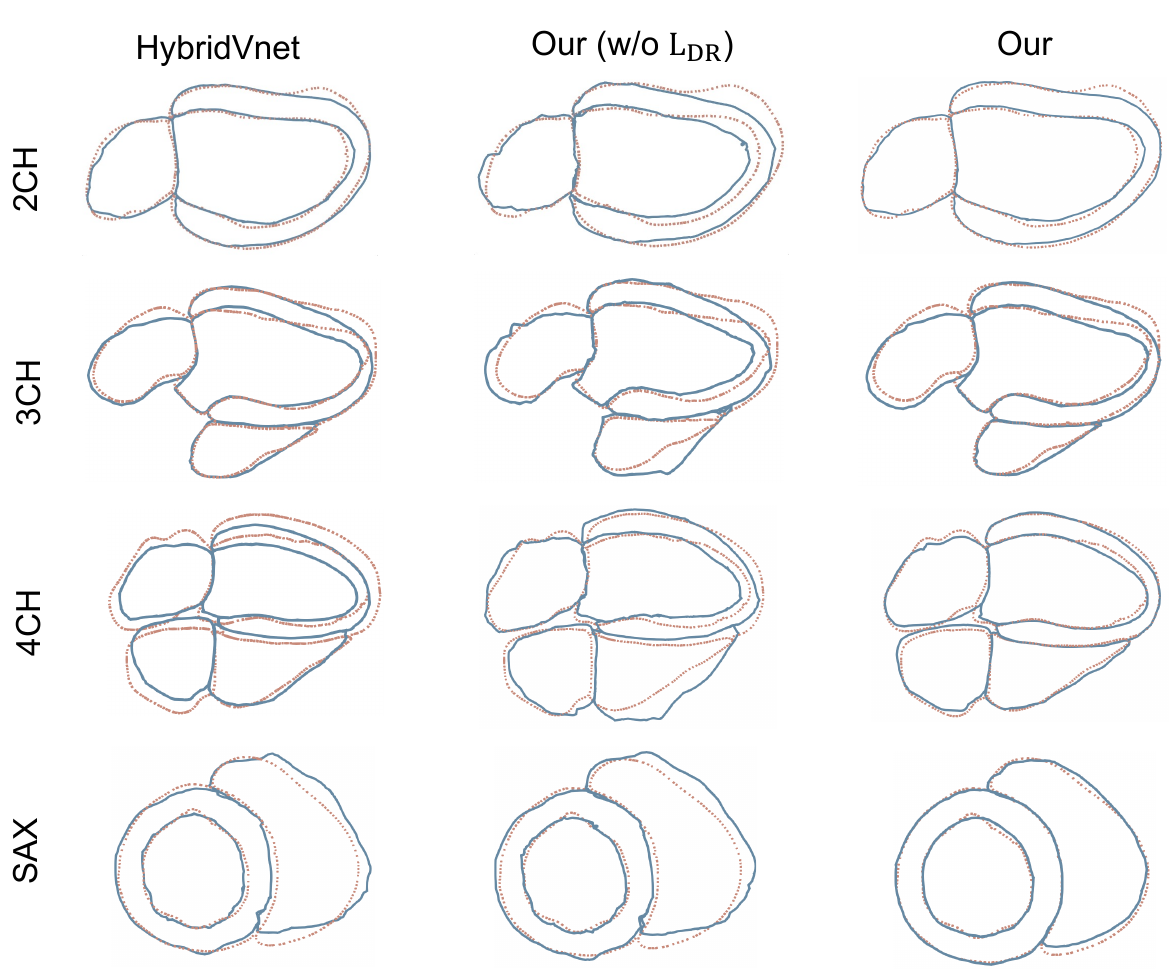}
  \caption{Qualitative comparison of multi-view contour alignment, where dash lines indicate ground truth and solid lines refer to estimated contours of different methods.}
\label{fig:result:boundary_overlay}
\end{figure}

The multi-scale temporal modeling block (w/o MST) increases MSE to 5.32 ± 2.02 mm\(^2\) and volumetric error to 24.7 ± 13.0\%, highlighting its importance for temporal coherence. 
This effect is further substantiated by the chamber volume curves presented in Fig.~\ref{fig:result:volume}. 
The full model closely tracks the ground-truth volume trajectory throughout the cardiac cycle, accurately capturing both the magnitude and phase of chamber contraction and relaxation. 
In contrast, the variant without multi-scale temporal modeling (w/o MST) exhibits attenuated volume changes and phase misalignment, particularly at end-systole, indicating that the absence of global-local temporal fusion leads to oversmoothed motion dynamics. 
These results confirm that the MST design, which jointly leverages sequence-level global motion and inter-frame consistency, is essential for modeling physiologically plausible cardiac motion.

The DR loss \(\mathcal{L}_{\mathrm{DR}}\) plays a critical role in enforcing contour alignment across imaging planes. 
As shown in Table~\ref{exp:tb:view_comparison}, removing \(\mathcal{L}_{\mathrm{DR}}\) (ours w/o DR) degrades boundary F-scores across all views, with the largest impact on long-axis views (2CH: from 65.47\% to 60.13\%; 3CH: from 65.71\% to 58.27\%; 4CH: from 68.24\% to 59.64\%). 
These gains are visually confirmed in Fig.~\ref{fig:result:boundary_overlay}, where the full model achieved precise contour overlap with ground-truth segmentations, whereas the variant without DR exhibits noticeable misalignments, particularly in regions with complex anatomy. 
This demonstrates that while vertex-wise MSE minimizes global distances, it does not penalize misalignment against specific 2D cuts, a limitation that \(\mathcal{L}_{\mathrm{DR}}\) explicitly addresses by rendering and comparing against 2D segmentations.

\subsection{External Validation on the Public Cine MRI Dataset} \label{exp:external_validation}


To assess the generalization capability of the proposed 4D whole-heart mesh reconstruction framework, we conducted external validation on the public CMR-MULTI-Fused cine MRI dataset released with the BAAI Cardiac Agent study~\citep{qu2026baai}. 
This dataset contains 105 cardiac MRI data collected under multi-center and missing-view (3CH view is not available) settings, thereby providing a challenging test bed for evaluating model robustness beyond the internal training distribution. 
We compared HybridVNet, the proposed model without LoRA adaptation, and the proposed model using both geometric reconstruction errors and temporal motion consistency metrics. 
Table~\ref{exp:tb:external_dataset_performance} presents the quantitative results on the external cine MRI dataset. 
LoRA adaptation consistently improved the proposed model across all evaluation metrics, reducing CD from 5.41~mm to 4.98~mm, HD from 17.33~mm to 16.33~mm, MAE from 3.44~mm to 2.69~mm, MSE from 20.33~mm$^2$ to 12.99~mm$^2$, and motion jitter $J_m$ from 1.94 to 1.82~mm/frame$^3$. 
Compared with HybridVNet, the proposed model achieved substantially lower MAE, MSE, and $J_m$, indicating improved point-wise reconstruction accuracy and temporal motion consistency. 
HybridVNet obtained slightly lower CD and HD, suggesting competitive global surface proximity; however, its higher point-wise errors and motion jitter indicate less accurate mesh correspondence and less stable temporal dynamics. 
Overall, these results demonstrate that the proposed framework generalizes effectively to external cine MRI data and that LoRA-based adaptation further improves robustness under cross-dataset distribution shift.

\begin{table}[t]
\centering
\caption{Performance on the external dataset for 4D whole-heart mesh reconstruction.}
{\scriptsize
\begin{tabular}{lccc}
\hline
Metric & HybridVNet & Ours (w/o LoRA)& Ours \\
\hline
CD (mm) $\downarrow$ & \textbf{4.88 $\pm$ 1.96} & 5.41 $\pm$ 0.17 & 4.98 $\pm$ 0.26\\
HD (mm) $\downarrow$ & \textbf{15.46 $\pm$ 3.67 } & 17.33 $\pm$ 0.84 & 16.33 $\pm$ 0.96 \\
MAE (mm) $\downarrow$ & 4.82 $\pm$ 1.04 & 3.44 $\pm$ 0.34  & \textbf{2.69 $\pm$ 0.39}\\
MSE (mm$^2$) $\downarrow$ & 25.61 $\pm$ 5.05 & 20.33 $\pm$ 3.87 & \textbf{12.99 $\pm$ 3.91}\\
${J}_{m}$ ($\mathrm{mm}/\mathrm{frame}^{3}$) $\downarrow$ & 2.90 $\pm$ 1.12 & 1.94 $\pm$ 1.02 & \textbf{1.82$\pm$ 1.88}\\
\hline
\end{tabular}}
\label{exp:tb:external_dataset_performance}
\end{table}

\subsection{Proof-of-Concept \textit{In-Silico} Electrophysiological Simulation} \label{exp:simulation} 

\begin{figure}[t]
  \centering
  \includegraphics[width=\linewidth]{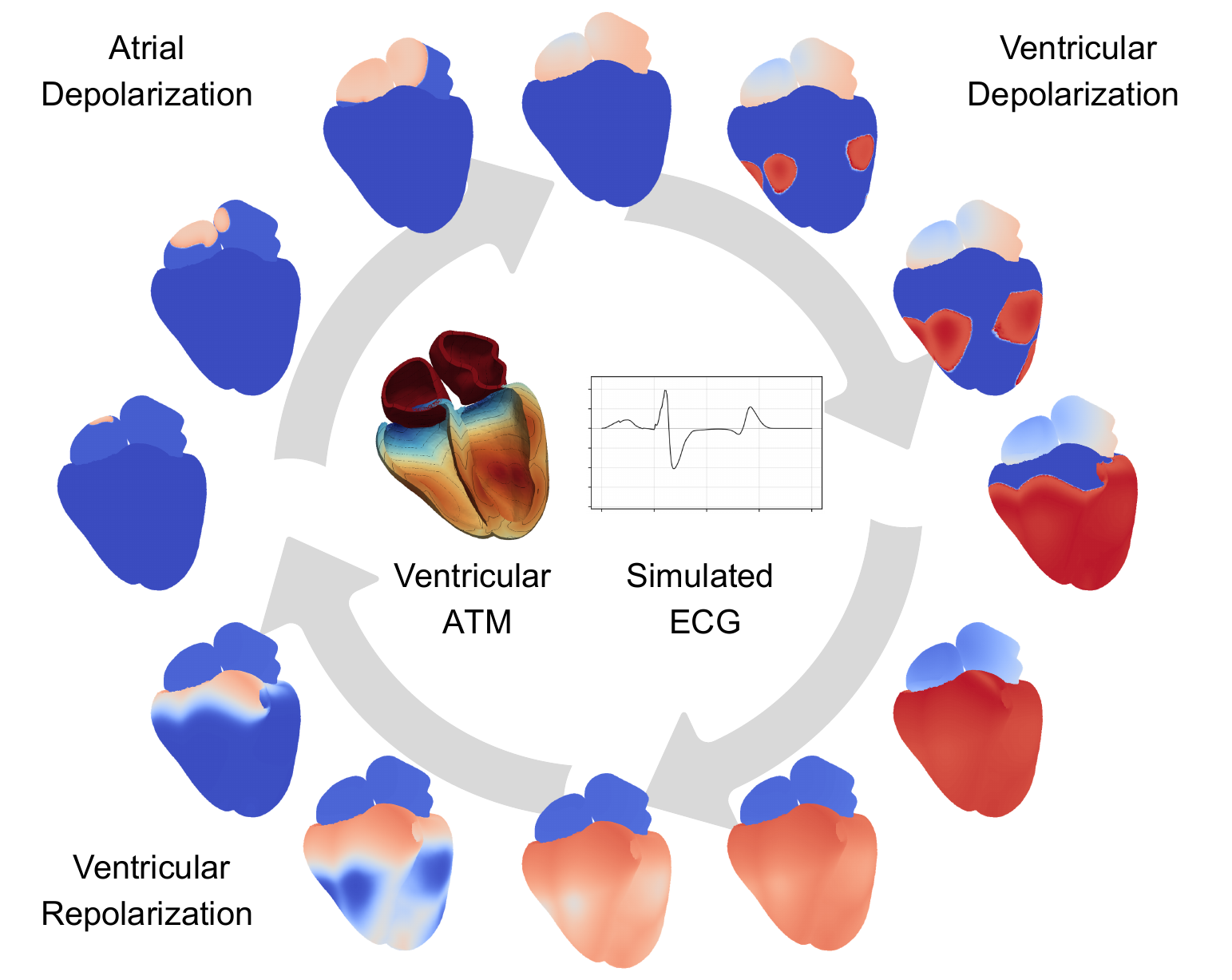}
  \caption{Whole-heart electrophysiological simulation over one cardiac cycle. The outer sequence shows transmembrane-potential propagation during atrial depolarization, ventricular depolarization, and ventricular repolarization. The center shows the ventricular activation-time map (ATM) and the simulated ECG.}
  \label{fig:ep_sim}
\end{figure}

To assess whether the reconstructed anatomy can support downstream computational modeling, we performed whole-heart EP simulation on a representative case. 
The reconstructed surface mesh was first converted into an upsampled volumetric mesh using the InSilicoHeartGen pipeline, where Cobiveco local coordinates, ventricular fibre orientation, and anatomy-specific anisotropy were assigned~\citep{journal/CMPB/doste2026,journal/MedIA/schuler2021}. 
The atria were modelled with isotropic conduction. 
For the ventricles, a simplified His-Purkinje system was incorporated by defining seven early activation sites, following~\citet{journal/CET/gerach2023}, together with a thin fast-conducting subendocardial layer whose conduction velocity was set to twice that of the surrounding myocardium. 
Both the His-Purkinje root locations and the subendocardial layer were defined in Cobiveco local coordinates~\citep{journal/MedIA/gillette2021}, while atrial activation sites were manually specified.
EP propagation was simulated using the monodomain model implemented in openCARP~\citep{journal/CMPB/plank2021}. 
The atria and ventricles were treated as separate physical regions with different ionic models: the Courtemanche model for the atria and the Ten Tusscher-Panfilov model for the ventricles. 
To approximate atrioventricular conduction delay, ventricular stimulation was initiated 200~ms after atrial stimulation. 
The torso was approximated as an infinite homogeneous volume conductor, from which a simulated ECG signal was computed. 
As shown in Fig.~\ref{fig:ep_sim}, the simulation reproduced the expected sequence of atrial depolarization, ventricular depolarization, and ventricular repolarization, together with a physiologically plausible ventricular activation-time map and ECG waveform. 
The ventricular activation pattern was qualitatively consistent with classical experimental observations~\citep{journal/Cir/durrer1970}. 
Although the current simulation uses generic EP parameters without subject-specific personalization or regional action-potential-duration heterogeneity, this proof-of-concept experiment suggests that the reconstructed whole-heart mesh can be converted into a simulation-ready anatomical substrate for downstream EP modeling.

\section{Discussion and Conclusion}

In this study, we have presented a fully automatic end-to-end framework for personalized 4D whole-heart mesh reconstruction from sparse multi-view 2D cine MRI. 
The proposed model achieved superior geometric accuracy compared to state-of-the-art methods, as demonstrated in Tables~\ref{exp:tb:substructure_comparison} and~\ref{exp:tb:comparison_study}. 
This performance gain can be attributed to three key contributions. 
First, the differentiable rendering loss grounded in the Beer-Lambert law enables anatomy-aware supervision from sparse 2D segmentations.
As shown in Table~\ref{exp:tb:view_comparison} and Fig.~\ref{fig:result:boundary_overlay}, removing this loss (ours w/o DR) substantially degraded boundary F-scores, particularly in long-axis views, confirming that vertex-wise MSE alone cannot enforce alignment against specific 2D cuts. 
Second, the MST modeling block, which fuses global cycle-level context with local inter-frame consistency, proved essential for physiologically plausible motion. 
The ablation study (Table~\ref{exp:tb:ablation_study}) shows that removing this block (w/o MST) increased volumetric error, and Fig.~\ref{fig:result:volume} further illustrates that the full model closely tracks ground-truth volume trajectories while the w/o MST variant exhibits attenuated contraction and phase misalignment. 
Third, unlike prior works that focus narrowly on biventricular myocardium \citep{conf/ICCV/yuan2023,journal/MedIA/laumer2025}, our method reconstructs complete four-chamber meshes. 
Notably, we observed that ventricular reconstructions consistently outperformed atrial ones, which we attribute to the fundamental sampling asymmetry in standard multi-view protocols: ventricles benefit from dense short-axis stacks while atria rely on only a single long-axis slice per view (see Section~\ref{exp:result:ours}). 
This disparity was further corroborated by the 2D contour analysis in Table~\ref{exp:tb:view_comparison}, where SAX views achieved the lowest mean contour distance due to dense sampling constraints. 
Finally,the proof-of-concept in-silico EP simulation (Section~\ref{exp:simulation}) suggests that the reconstructed whole-heart meshes can serve as simulation-ready anatomical substrates, supporting the feasibility of using routine cine MRI for future CDT modeling.

Nevertheless, three limitations of this work should be acknowledged. 
First, our method relies on accurate 2D segmentations from cine MRI as supervision. 
While we employed an automated pipeline with manual refinement, segmentation errors can propagate to the reconstructed meshes. 
As shown in Table~\ref{exp:tb:view_comparison}, the boundary F-scores for long-axis views remain lower than ideal, reflecting residual segmentation uncertainty. 
Second, the current framework reconstructs whole-heart geometry but does not incorporate tissue property estimation (e.g., myocardial stiffness, conductivity) or biomechanical constraints, which are essential for downstream EP or electromechanical simulations. 
Third, our method has only been tested on post-MI patients. 
While this cohort represents a clinically relevant population with diverse infarct patterns, the generalizability to other pathologies (e.g., cardiomyopathy, valvular disease, atrial fibrillation) and to healthy subjects remains to be validated. 
Additionally, as shown in Fig.~\ref{fig:result:volume}, the predicted LV volume change appeared attenuated compared to ground truth. 
The reduced LV volume variation in the predicted curves may reflect the difficulty of modeling subtle, pathology-altered contraction patterns in post-MI patients. 
Although post-MI physiology is characterized by impaired contractility, the remaining discrepancy with the reference curves indicates that further improvement is needed for accurately capturing motion amplitude.

In our future work, we plan to address these limitations and extend the framework toward a fully integrated CDT platform. 
First, we will incorporate self-supervised or weakly supervised learning strategies to reduce dependence on manual segmentations. 
Second, we aim to integrate biomechanical priors into the reconstruction pipeline, enabling direct estimation of tissue properties and patient-specific electromechanical parameters from the same sparse cine MRI inputs. 
Third, we will evaluate the proposed method on a larger multi-center population across different scanner vendors, imaging protocols, and disease cohorts (including healthy controls) to thoroughly investigate its generalization ability. 
On this larger dataset, we can perform a comprehensive sensitivity analysis to quantitatively investigate the impact of sampling variability and slice misalignment on reconstruction accuracy. 
Fourth, we plan to extend the temporal modeling to handle variable-length and arrhythmic cycles using sequence-to-sequence models or neural ordinary differential equations, broadening clinical applicability to patients with rhythm disorders. 
Additionally, we will explore a more detailed patient-specific whole-heart-torso modeling framework, incorporating cardiac motion and distinct anatomical compartments (atria, ventricles, great vessels, pericardium) to enhance the fidelity of downstream simulations. 
Finally, we will work toward an end-to-end training scheme that jointly optimizes segmentation, reconstruction, and temporal modeling, potentially via differentiable graph networks or implicit neural representations, to further streamline the pipeline and reduce error propagation. 
In conclusion, this work demonstrates the feasibility of automatic 4D whole-heart mesh reconstruction from routine clinical cine MRI, achieving both geometric accuracy and temporal coherence, paving the way for future research to enable large-scale, real-time CDT applications.

\bibliographystyle{model2-names}
\biboptions{authoryear}
\bibliography{A_refs}

\end{document}